%% file: root.tex
\pdfoutput=1
\documentclass[letterpaper,10pt,conference]{ieeeconf}
\IEEEoverridecommandlockouts
\usepackage{iftex}
\ifPDFTeX
  \usepackage[T1]{fontenc}
  \usepackage{tgtermes,tgheros,tgcursor}
  \usepackage{microtype}
\else
  \usepackage{fontspec}
\fi
\usepackage{graphicx}
\usepackage{float}
\usepackage{amsmath,amssymb}
\usepackage{pifont}
\usepackage{fix-cm}
\usepackage{array,booktabs}
\usepackage{cite}
\usepackage[nocheckfootnote]{flushend}
\usepackage{placeins}
\usepackage{xcolor}
\usepackage{url}
\usepackage{tikz}
\usepackage{pgfplots}
\pgfplotsset{compat=1.18}
\usepackage[bookmarks=false,hidelinks]{hyperref}
\input{layout}
\usepackage{etoolbox}

\hypersetup{pdftitle={Brick But Agile: A Compact, Modular Anthropomorphic LEGO Dexterous Hand for Education and Research},pdfauthor={Renjun Gao, Cheng Li, Yuhang Hu, Xiaolei Ren, Guyue Zhou}}
\patchcmd{\artwork}{{\scriptsize Artwork placeholder\quad /\quad height #3}}{}{}{}

\renewcommand{\widefigureheader}[5]{%
 \begin{figure*}[!t]\centering
  \artwork{#1}{\textwidth}{#2}{#3}%
  \caption{#4}\label{#5}%
 \end{figure*}
}

\title{\LARGE \bf
Brick But Agile: A Compact, Modular Anthropomorphic LEGO\textsuperscript{\textregistered} Dexterous Hand for Education and Research
}
\author{Renjun~Gao$^{1,2,\dagger,\ddagger}$, Cheng~Li$^{3,\dagger}$, Yuhang~Hu$^{4}$,
Xiaolei~Ren$^{2}$, and Guyue~Zhou$^{1,\dagger}$\\[0.35em]
{\small $^{1}$Institute for AI Industry Research (AIR), Tsinghua University}\quad
{\small $^{2}$Macau University of Science and Technology}\\
{\small $^{3}$The Hong Kong University of Science and Technology}\quad
{\small $^{4}$Independent Researcher}\\[0.15em]
{\footnotesize $^{\dagger}$Corresponding authors: \href{mailto:renjgao@gmail.com}{renjgao@gmail.com}, \href{mailto:clieo@connect.ust.hk}{clieo@connect.ust.hk}, \href{mailto:zhouguyue@air.tsinghua.edu.cn}{zhouguyue@air.tsinghua.edu.cn}\qquad $^{\ddagger}$Project lead}
}
\bbateaser{Brick But Agile (BBA). Left: instruction-guided LEGO assembly. Center: vision-based teleoperation and tool use. Right: exploded architecture showing detachable finger and enclosure modules, tendon routing, motors, and control hubs.}
\begin{document}
\maketitle
\thispagestyle{empty}
\pagestyle{empty}
\raggedbottom
\bstctlcite{IEEEexample:BSTcontrol}

\begin{abstract}
Dexterous hands enable versatile robotic manipulation, yet custom fabrication can limit access to these platforms. LEGO's readily available parts and reversible connections simplify reproduction and modification. We present Brick But Agile (BBA), the first LEGO dexterous robotic hand: a modular, human-scale five-finger platform assembled from standard LEGO bricks without custom-fabricated parts. Eight active inputs coordinate motion about eighteen joint axes: four for independent non-thumb finger flexion, three for the thumb, and one for human-like finger spreading. Within each non-thumb finger, passive distal coupling and elastic return provide contact adaptation. A user-friendly graphical interface supports manual control, motion sequencing, vision-based teleoperation, and synchronized demonstration capture, facilitating hands-on robotics education. We also provide a MuJoCo simulation model that explicitly represents the tendon-driven transmission. Experiments demonstrate BBA's versatile grasping capabilities: it reproduces all 33 GRASP taxonomy grasp types and scores 10/10 on the Kapandji test. Despite its interlocking-brick construction, BBA produces a mean maximum fingertip force of 3.30~N and supports a 2.35~kg static payload. BBA also enables reliable vision-based teleoperation. Editable assembly models and a shared hardware--simulation command interface make BBA an integrated platform for dexterous-manipulation research and hands-on robotics education. \mbox{Visit \href{https://bba.renjun.one}{\textcolor[HTML]{0066FF}{the project webpage}} for details.}
\end{abstract}

\section{Introduction}

Human demonstrations, teleoperation, and robot learning are expanding the capabilities of multi-fingered hands~\cite{wang2024dexcap,iyer2024openteach}. Open-source platforms such as RUKA, ORCA, Aero Hand Open, and LEAP support this research, but require 3D-printed or custom-fabricated parts~\cite{zorin2025ruka,christoph2025orca,wang2026aero,shaw2023leap}. For schools, individual makers, and researchers without fabrication facilities, practical access also depends on how readily these mechanisms can be assembled and modified.

The LEGO System offers a familiar construction platform for hands-on robotics, building on decades of LEGO Technic and programmable LEGO robotics in education~\cite{danahy2014lego,resnick1996bricks}. Standard LEGO bricks and Technic elements, together with illustrated instructions, allow mechanisms to be built without custom fabrication. Reversible connections and the My Own Creation (MOC) ecosystem support inspection, rebuilding, customization, and sharing of editable designs~\cite{bricklinkstudiogallery}.

\begin{figure}[!t]
\centering
\includegraphics[width=\columnwidth]{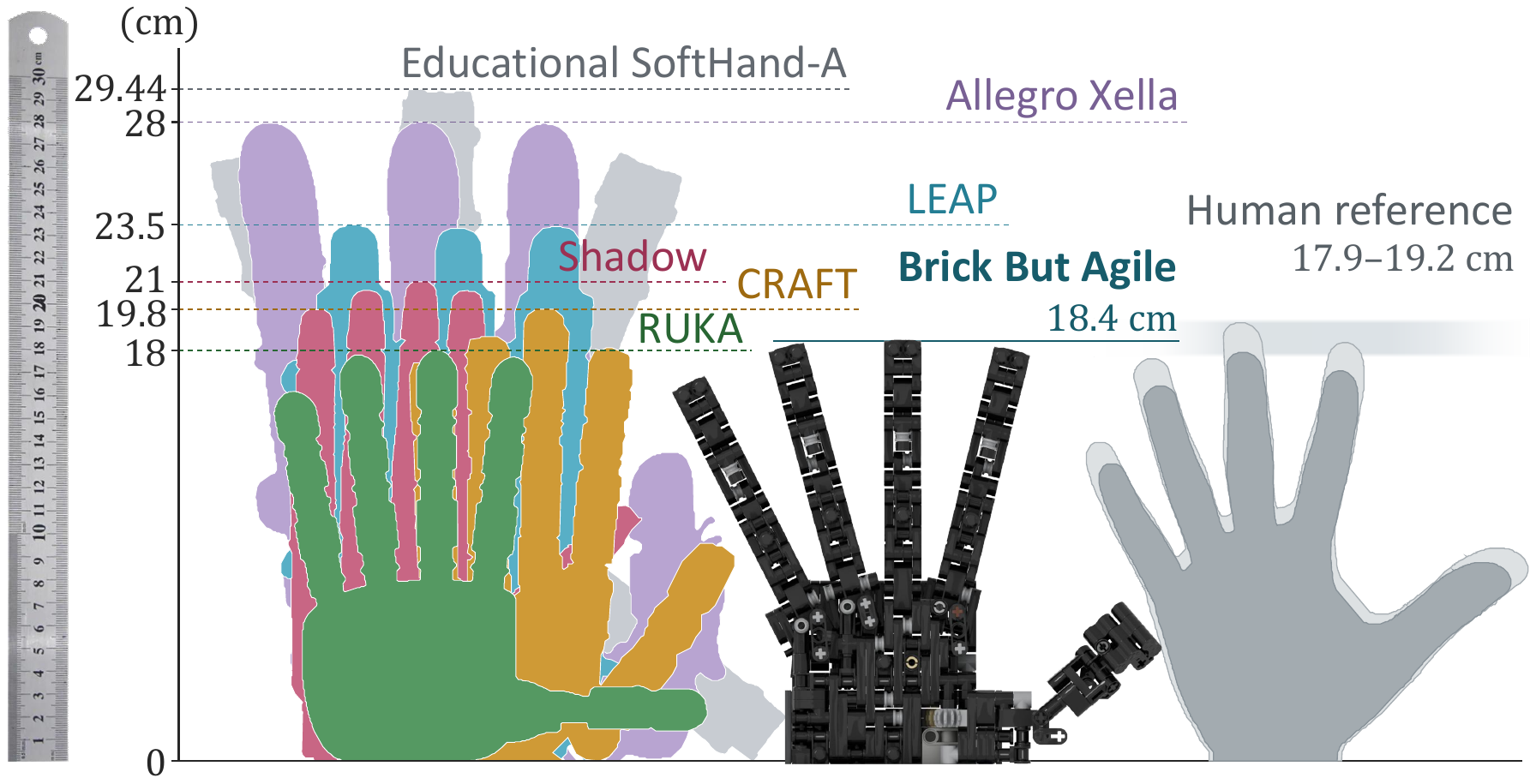}
\caption{Hand-size comparison. Reference sizes for Allegro Xella, LEAP, Shadow, and RUKA are from~\cite{zorin2025ruka}; CRAFT from~\cite{lin2026craft}. Educational SoftHand-A is estimated from its published LEGO assembly~\cite{lepora2025educational}. BBA is 18.4~cm long; the human reference spans 17.9--19.2~cm~\cite{webb1978anthropometric}.}
\label{fig:morphology}
\end{figure}

However, accessible construction must still preserve the geometry and motion required for dexterous manipulation. Palm-to-finger proportions, phalanx lengths, and joint arrangements govern fingertip reach and contact orientation and provide a geometric basis for retargeting human motion~\cite{zorin2025ruka,christoph2025orca,piazza2019century}. Achieving these requirements at human scale is particularly difficult with LEGO because fixed dimensions, hole spacings, and connection geometries restrict mechanism design to discrete combinations. As the hand becomes smaller, joint supports and transmission elements must therefore be rearranged rather than simply scaled down. Educational SoftHand-A illustrates this trade-off: to simplify construction, it adopts four digits with $145\times30$~mm finger modules~\cite{lepora2025educational}, whereas our design retains five fingers with $104\times16$~mm finger modules and an overall hand length of 18.4~cm, within the 17.9--19.2~cm human reference range shown in Fig.~\ref{fig:morphology}. Despite these discrete-part constraints, this compact morphology preserves selective finger flexion, thumb opposition, active finger spread, and passive contact adaptation.

We present Brick But Agile (BBA), the first LEGO dexterous robotic hand, pairing a human-scale five-finger morphology with selective grasp control, vision-based teleoperation, and tendon-driven simulation (Fig.~\ref{fig:teaser}). Its tendon--gear--linkage transmission coordinates eighteen joint axes through eight active inputs: four for independent non-thumb finger flexion, three for thumb positioning, and one for shared abduction and adduction (Table~\ref{tab:architecture}). These inputs control grasp preshaping, while staged closure, passive distal-joint coupling, and elastic return coordinate each finger's response to contact. Detachable finger modules exploit LEGO's reversible connections for rapid replacement and reconfiguration (Fig.~\ref{fig:teaser}).

The graphical interface provides manual control, pose storage, motion sequencing, and vision-based teleoperation. Hardware and simulation share the same eight-input command space, with synchronized demonstration capture, logging, and replay for reproducible experiments. Together with the editable construction, these tools enable a build--modify--control--observe workflow for hands-on study of robotic mechanisms and grasping.

The main contributions are as follows:
\begin{enumerate}
    \item \textbf{A compact, modular, human-scale LEGO dexterous hand.}
    Within LEGO's discrete dimensional and connection constraints, we realize an 18.4~cm five-finger hand with four detachable $104\times16$~mm finger modules, leveraging reversible standard-brick connections for fabrication-free assembly and reconfiguration.

    \item \textbf{An eight-input hybrid transmission and anthropomorphic mechanisms for versatile grasping.}
    Our tendon--gear--linkage architecture coordinates 18 joint axes using 8 LEGO motors: 4 for independent finger flexion, 3 for the thumb, and 1 for finger abduction/adduction. Underactuated fingers combine joint-dependent moment arms, passive PIP--DIP coupling, and elastic return for staged contact adaptation, while geared CMC opposition, an inclined IP axis, and a three-branch palm linkage orient the thumb and generate human-like spread. Experimental results, including 33/33 in the GRASP taxonomy~\cite{feix2016} and 10/10 on the Kapandji test~\cite{kapandji1986}, confirm the versatile performance of the hand.

    \item \textbf{An integrated platform for hands-on robotics education and dexterous-manipulation research.}
    A user-friendly graphical interface provides manual control and motion sequencing, while our software stack supports reliable vision-based teleoperation and data collection for robot-learning applications. A MuJoCo model explicitly simulates the tendon-driven transmission. Together, these tools enable a build--modify--control--observe workflow.
\end{enumerate}

\section{Related Work}

\begin{table}[!t]
\caption{Hand architecture and grasp coverage}
\label{tab:architecture}
\centering
\begingroup
\footnotesize
\setlength{\tabcolsep}{1pt}
\renewcommand{\arraystretch}{1.08}

\begin{tabular*}{\columnwidth}{@{\extracolsep{\fill}}C{21mm} C{8.2mm} C{9.5mm} C{10.2mm} C{13.3mm} C{9.5mm} C{9.3mm}@{}}
\toprule
\textbf{Hand} &
\textbf{Digits} &
\textbf{DOF/\newline DOA} &
\textbf{Act.} &
\textbf{Active\newline spread} &
\textbf{Grasps\newline (n/33)} &
\textbf{LEGO} \\
\midrule

LEAP~\cite{shaw2023leap}
& 4 & 16/16 & D & Individual & --- & \ding{55} \\

Shadow~\cite{shadow2024spec}
& 5 & 22/18\textsuperscript{\dag} & T & Individual & --- & \ding{55} \\

Allegro~\cite{wonikAllegroV4}
& 4 & 16/16 & D & Individual & --- & \ding{55} \\

RUKA~\cite{zorin2025ruka}
& 5 & 15/11 & T & None & 29/33 & \ding{55} \\

Ruka-v2~\cite{liu2026rukav2}
& 5 & 18/14\textsuperscript{\dag} & T & I/R/L indiv. & --- & \ding{55} \\

ORCA~\cite{christoph2025orca}
& 5 & 16/16\textsuperscript{\dag} & T & Individual & --- & \ding{55} \\

CRAFT~\cite{lin2026craft}
& 5 & 20/15 & T+L & Individual & 33/33 & \ding{55} \\

DexLink~\cite{wu2026dexlink}
& 5 & 20/16 & L+W & Individual & 33/33 & \ding{55} \\

Educational\newline
\mbox{SoftHand-A~\cite{lepora2025educational}}
& 4 & 12/2 & T+C & None & --- & \ding{51} \\

\midrule

\textbf{Brick But Agile}
& \textbf{5}
& \textbf{18/8}
& \textbf{T+D+L}
& \textbf{Shared}
& \textbf{33/33}
& \ding{51} \\

\bottomrule
\end{tabular*}

\par\vspace{3pt}
\parbox{\columnwidth}{\fontsize{7.3}{8.5}\selectfont
DOF/DOA: degrees of freedom/actuation. Act.: D, direct; T, tendon; L, linkage; W, worm gear; C, clutch gear. \textsuperscript{\dag}Wrist excluded. ---: not reported.}

\endgroup
\end{table}

\subsection{Actuation Allocation in Anthropomorphic Hands}

Actuator placement and transmission design determine how independent motion control is integrated within an anthropomorphic hand~\cite{piazza2019century} (Table~\ref{tab:architecture}). Direct-drive designs such as LEAP place actuators at or near the finger joints, whereas tendon-driven RUKA and ORCA relocate actuation proximally~\cite{shaw2023leap,zorin2025ruka,christoph2025orca}. This frees space in the hand for joint mechanisms, but introduces tendon routing, tensioning, and calibration challenges.

At lower actuation dimensionality, the Pisa/IIT SoftHand and BRL/Pisa/IIT SoftHand use adaptive synergies to coordinate whole-hand closure and contact adaptation with a single actuator~\cite{catalano2014,li2022brl}. Aero Hand Open instead gives each non-thumb finger its own tendon input while mechanically coupling motion within the finger~\cite{wang2026aero}. These approaches illustrate the design choice between globally coordinated closure and selective finger participation.

Finger abduction and adduction extend grasp preshaping to lateral contact placement. RUKA reproduces 29 of the 33 GRASP taxonomy types without active finger abduction~\cite{zorin2025ruka}. CRAFT, which includes active side-to-side finger motion, reports 33/33 and highlights lateral motion for inter-finger grasps and tool handling~\cite{lin2026craft}. Ruka-v2 explicitly adds controlled abduction/adduction for thin-object grasping and in-hand reorientation~\cite{liu2026rukav2}. A lower-actuator alternative is demonstrated by iCub, where one motor coordinates abduction/adduction of the index, ring, and little fingers~\cite{schmitz2010}. BBA realizes shared lateral actuation through a three-branch palm linkage while retaining independent non-thumb flexion, allowing finger spread to be commanded separately from flexion.

\subsection{Reproducible Construction and LEGO Robotics}

Beyond 3D-printed open hardware, the LEGO ecosystem has a long history in accessible robotics: Technic supplies reusable mechanical elements, while programmable platforms such as MINDSTORMS combine them with motors, sensors, and controllers~\cite{resnick1996bricks,danahy2014lego}. Studio and BrickLink further enable editable digital assemblies and standardized part sourcing~\cite{bricklinkstudiogallery}.

Educational SoftHand-A brings two-actuator synergy grasping into LEGO construction~\cite{lepora2025educational}; BBA advances LEGO hand design to a human-scale five-finger architecture with selective finger flexion, three-axis thumb actuation, shared finger spread, teleoperation, and simulation.

\subsection{Transmission Calibration and Teleoperation}

Teleoperation of tendon-driven hands requires two mappings: human motion must be retargeted to feasible robot configurations, and desired robot motion must be converted to actuator commands. Recent tendon-driven hands address this second stage through learned, calibrated, or model-based transmission models~\cite{zorin2025ruka,christoph2025orca,liu2026rukav2,wang2026aero}.

OPEN TEACH and DexCap demonstrate accessible teleoperation and portable human-demonstration capture for dexterous manipulation~\cite{iyer2024openteach,wang2024dexcap}. BBA combines calibrated execution, vision-based demonstration capture, and tendon-driven simulation through a shared eight-input command space.

\begin{figure*}[t]
\centering
\includegraphics[width=\textwidth]{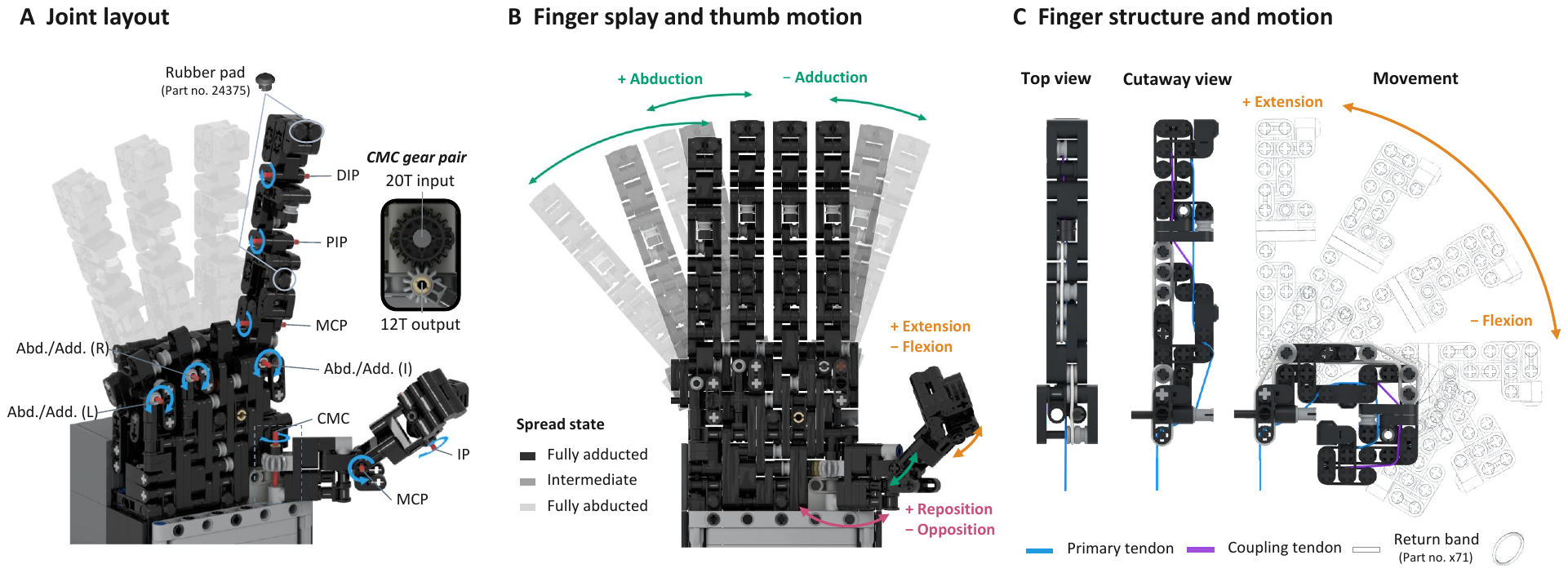}
\caption{BBA hand mechanisms. (A) Joint axes and CMC gear pair. (B) Finger spread and thumb motions. (C) Finger structure, tendon routing, and flexion. Blue, purple, and white denote the primary tendon, PIP--DIP coupling tendon, and return bands.}
\label{fig:mechanisms}
\end{figure*}
\section{Mechanical and Kinematic Design}
\label{sec:hardwarealias}
\subsection{Overall Structure and Actuation}
BBA consists of four non-thumb fingers, an opposable thumb, a palm, and a forearm actuation module. Each non-thumb finger module measures $104\times16$~mm, the palm is 80~mm long, and the total hand length is 184~mm. The extended hand is four LEGO units thick, where one LEGO unit denotes the nominal 8~mm stud pitch. The hand has 18 joint axes and 8 motors: 4 for independent non-thumb finger flexion, 3 for thumb motion, and 1 for shared finger abduction and adduction (Fig.~\ref{fig:mechanisms}). All motors are housed in the forearm, leaving space within the fingers for joints and transmission elements. In the fully adducted reference pose, the non-thumb fingers are parallel with zero spread angle and a nominal half-LEGO-unit gap between adjacent finger modules.

LEGO plastic tubes (Hose, Rigid 3mm D., part numbers in the 75c series) route the tendons between the forearm and palm. Because the tubes can bend along narrow internal passages, their smooth inner walls provide continuous low-friction guidance while keeping the tendons clear of structural edges.

\begin{figure}[t]
\centering
\includegraphics[width=\columnwidth]{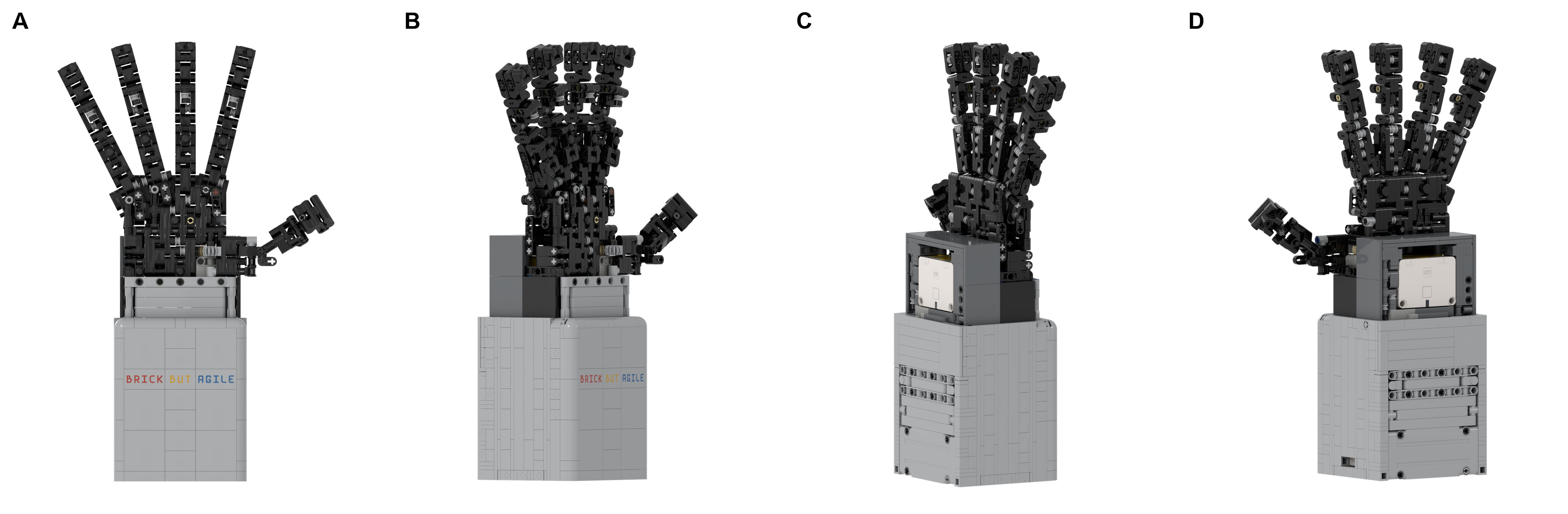}
\caption{Palmar, oblique palmar, oblique dorsal, and dorsal rendered views (A--D) of BBA's removable System-brick enclosure around the Technic mechanisms.}
\label{fig:enclosure_views}
\end{figure}

BBA combines Technic elements for the mechanisms and transmissions with removable LEGO System-brick modules for the external enclosure. Stud-concealing construction integrates the actuation and control hardware into a coherent exterior while preserving access to the internal modules (Fig.~\ref{fig:enclosure_views}).

\begin{table}[t]

\caption{Range of Motion Compared with the Human Hand}

\label{tab:rom}

\centering

\begingroup

\fontsize{8.5}{10.1}\selectfont

\setlength{\tabcolsep}{3pt}

\renewcommand{\arraystretch}{1.08}

\begin{tabular*}{\columnwidth}{@{\extracolsep{\fill}}C{19mm} C{23mm} C{20mm} C{17mm}@{}}
\toprule
\textbf{Type} & \textbf{Motion} & \textbf{Human ROM} & \textbf{BBA ROM} \\
\midrule
\textbf{Finger} & MCP flex. & $100^\circ$ & $90^\circ$ \\
 & PIP flex. & $105^\circ$ & $90^\circ$ \\
 & DIP flex. & $85^\circ$ & $90^\circ$ \\
\midrule
\textbf{Thumb} & CMC & --- & $170^\circ$ \\
 & MCP Abd./Add. & $32^\circ$\textsuperscript{*} & $90^\circ$ \\
 & IP flex. & $73^\circ$ & $90^\circ$ \\
\midrule
\textbf{Abd./Add.} & Index & $16.8\pm2.6^\circ$ & $26.3^\circ$ \\
 & Middle & Reference & Fixed \\
 & Ring & $14.8\pm2.5^\circ$ & $25.5^\circ$ \\
 & Little & $32.8\pm3.8^\circ$ & $40.9^\circ$ \\
\bottomrule
\end{tabular*}

\par\vspace{3pt}

\parbox{\columnwidth}{\fontsize{7.3}{8.5}\selectfont

ROM: range of motion; flex.: flexion; Abd./Add.: abduction/adduction. Human finger-flexion values and the thumb-IP flexion endpoint are from the full active ROM data in~\cite{hume1990rom}. Human finger-abduction values are the mean maximum right-hand angles measured relative to the middle finger~\cite{chatzioglou2024hand}. \textsuperscript{*}Human thumb MCP Abd./Add. is the $32^\circ$ motion range measured during opposition--reposition in 18 healthy subjects~\cite{hamann2014thumb}. BBA finger-spread angles use the same middle-finger reference. ---: not specified.}

\endgroup

\end{table}

All eight actuators are LEGO Powered Up Technic Large Motors with integrated rotation sensors (part number~88013). A SPIKE Prime Technic Large Hub (part number~45601) and a SPIKE Essential Technic Small Hub (part number~45609) provide six and two motor ports, respectively; both run Pybricks firmware. Standard LEGO components therefore provide the structure, actuation, and control hardware throughout the platform.

The finger, thumb, spread, and forearm actuation modules can be detached and replaced individually, providing direct access to tendon routing and return bands for inspection and adjustment. Step-by-step instructions guide assembly, while an editable model in BrickLink Studio~\cite{studio} preserves part identifiers and assembly relationships for reproduction and modification.

\subsection{Compact Adaptive Fingers}
The four non-thumb fingers share an underactuated design. A single primary tendon drives the metacarpophalangeal (MCP), proximal interphalangeal (PIP), and distal interphalangeal (DIP) flexion joints. The tendon's offsets from the joint axes define joint-dependent moment arms through which tension generates flexion torque (Fig.~\ref{fig:mechanisms}(C)). A separate passive tendon couples PIP and DIP motion at an angular ratio of approximately $1{:}1$. LEGO $2\times2$ rubber bands (part number~x71) provide extension. The bands sit in the central grooves of half bushes (part number~4265c), spanning 4 and 5 LEGO units at the MCP and PIP joints, respectively. Because the two-LEGO-unit finger width comprises four half-LEGO-unit layers, each band sits on an inner off-center layer rather than the midplane, introducing a slight lateral offset. Together, the return-band configurations and mechanical stops produce a staged closing sequence. During free motion, the MCP joint first reaches its stop at approximately $90^\circ$; further tendon take-up then primarily flexes the PIP and DIP joints. When contact constrains a finger segment, the remaining joints can continue closing to conform to the object.

Each finger is two LEGO units (16~mm) wide, where fixed part dimensions and hole spacings tightly constrain joint clearances and the placement of supports, tendon guides, and return bands. Along the 104~mm module length, this compact section accommodates all three flexion joints, internal tendon guidance, distal coupling, and elastic return. No LEGO connection in the finger is oriented along its longitudinal axis, so pulling along the finger does not separate the parts. Where the tendon path changes direction, the tendon passes around cylindrical LEGO elements or over the smooth curved surfaces of liftarms to reduce friction. The distal and proximal phalanges are constructed to accommodate standard LEGO rubber pads (part number~24375), which increase contact friction and compliance; at full flexion, the two pads meet. Fig.~\ref{fig:mechanisms}(C) shows the finger's top view, internal tendon routing, and flexion sequence.

\subsection{Three-Axis Thumb with Geared Opposition}
The thumb has three separately actuated revolute joints. The carpometacarpal (CMC) joint produces opposition toward the palm, the MCP joint provides abduction and adduction, and the interphalangeal (IP) joint controls distal flexion and extension (Fig.~\ref{fig:mechanisms}(B)). The MCP axis is perpendicular to the CMC axis, and the IP axis is inclined $60^\circ$ toward the palm relative to the MCP axis (Fig.~\ref{fig:mechanisms}(A)). This inclination is chosen to direct distal closure toward the opposing fingertips, following the functional role of oblique thumb axes in contact-surface orientation~\cite{chalon2010}. Together, these three axes position the thumb and orient its distal pad relative to the opposing fingers.

A 20-tooth input gear drives the compact 12-tooth integral-axle output gear (part~24014; Fig.~\ref{fig:mechanisms}(A)) with a $20{:}12$ tooth ratio. Within the four-LEGO-unit hand thickness and the fixed CMC axis arrangement, part~24014 is the only standard LEGO gear that satisfies the required clearance and attachment geometry. Direct gearing provides a fixed motor-to-joint relationship for thumb opposition, while separate tendons drive the MCP and IP joints. Table~\ref{tab:rom} summarizes the joint ranges of motion.

\subsection{Continuously Adjustable Finger Abduction and Adduction}
\begin{figure}[!t]
\centering
\includegraphics[width=\columnwidth]{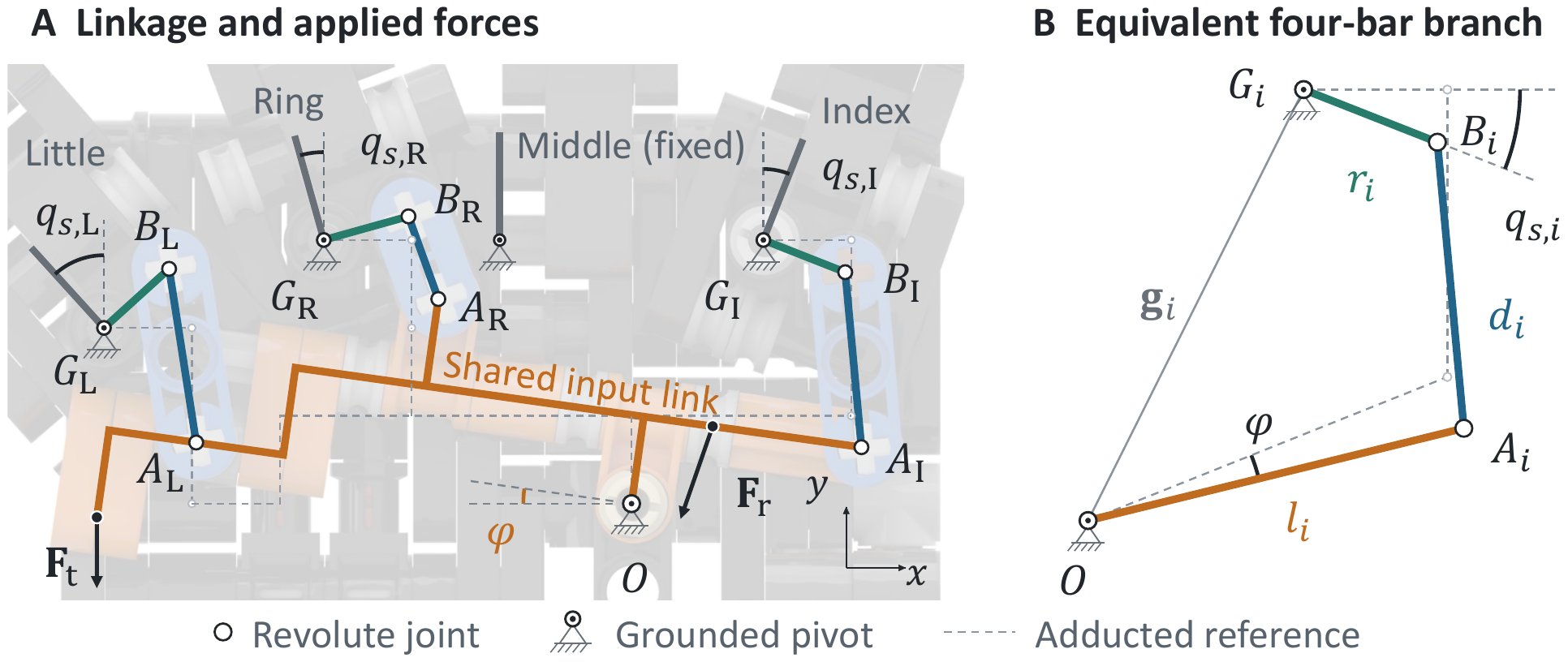}
\caption{Shared abduction/adduction linkage. (A) Linkage layout and applied forces. $\mathbf{F}_{\mathrm t}$ is the tendon tension, and $\mathbf{F}_{\mathrm r}$ is the resultant force of the two return bands. (B) An equivalent four-bar model of one branch. Orange, blue, and green indicate the input link, coupler, and finger-base rocker, respectively. Gray dashed lines show the fully adducted reference configuration.}
\label{fig:spread_linkage}
\end{figure}%
To retain active finger spread without separate abduction motors, we use a single motor to drive a three-branch linkage that coordinates abduction and adduction of the index, ring, and little fingers; the middle finger remains laterally fixed (Fig.~\ref{fig:mechanisms}(B)). Tendon tension $\mathbf{F}_{\mathrm t}$ acts on the extended end of a shared input link, rotating it about the fixed palm pivot $O$ to adduct the three fingers (Fig.~\ref{fig:spread_linkage}(A)). The resultant force $\mathbf{F}_{\mathrm r}$ from two return bands produces an opposing moment, spreading the fingers as the tendon is released. The branch geometry moves the index finger in the opposite lateral direction to the ring and little fingers, with greater abduction at the little finger. The three movable finger-base joints are designed so that, as the fingers spread, the flexion tendon does not slip laterally and the MCP return band remains clear of it.

The half-LEGO-unit gap maintains a human-like separation between adjacent digits while providing clearance during finger flexion. It also accommodates small lateral misalignments caused by LEGO assembly tolerances and the asymmetric return-band mounting within the two-unit-wide finger modules.

Each branch connects the shared input link to a finger-base rocker through a coupler $A_iB_i$, forming the equivalent four-bar linkage in Fig.~\ref{fig:spread_linkage}(B). The subscript $i\in\{I,R,L\}$ denotes the index, ring, and little fingers. Points $A_i$ and $B_i$ rotate about $O$ and the fixed finger-base pivot $G_i$, respectively. The input arm, coupler, and finger-base rocker have lengths $l_i$, $d_i$, and $r_i$, and $\mathbf{g}_i=\overrightarrow{OG_i}$ is the fixed base vector. Link lengths are measured between pivot centers in LEGO units; the base-pivot offsets and zero-pose link orientations define the assembled geometry.

In a palm-plane coordinate frame with origin $O$, define $\mathbf{a}_i(\varphi)=\overrightarrow{OA_i}$ (with $\lVert\mathbf{a}_i\rVert=l_i$) and $\mathbf{b}_i(q_{s,i})=\mathbf{g}_i+\overrightarrow{G_iB_i}$ (with $\lVert\overrightarrow{G_iB_i}\rVert=r_i$). The shared input angle $\varphi$ and finger abduction angles $q_{s,i}$ are measured from the adducted reference configuration. In the nominal planar model, the fixed coupler length imposes
\begin{equation}
\left\lVert \mathbf{a}_i(\varphi)-\mathbf{b}_i(q_{s,i}) \right\rVert^2=d_i^2,
\qquad i\in\{I,R,L\}.
\label{eq:spread}
\end{equation}

For a given assembly, \eqref{eq:spread} relates input-link rotation to the three coordinated abduction angles. The link lengths are chosen in discrete LEGO units, within palm-clearance limits, to emulate human hand opening: at approximately 75\% of the input link's angular travel, the assembled hand simultaneously reaches $15.6^\circ$, $16.8^\circ$, and $30.7^\circ$ for the index, ring, and little fingers, respectively. These values closely match the reported mean maximum right-hand abduction angles relative to the middle finger~\cite{chatzioglou2024hand}, while full input travel reaches the $26.3^\circ$, $25.5^\circ$, and $40.9^\circ$ maxima in Table~\ref{tab:rom}.

\begin{figure*}[!t]
\centering
\includegraphics[width=1\textwidth]{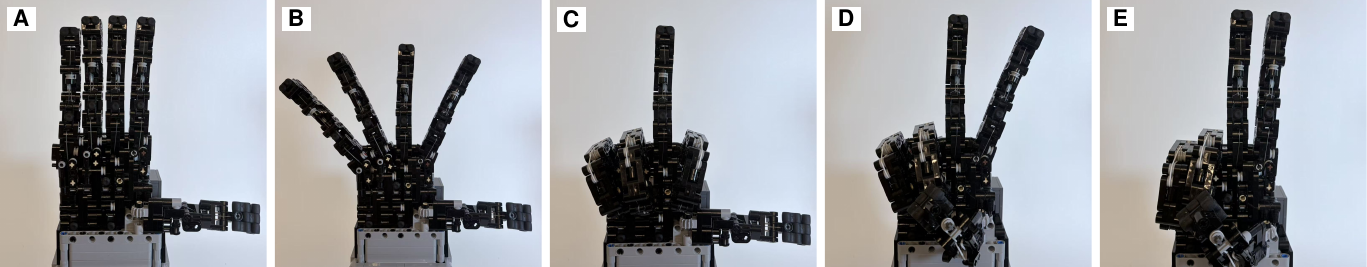}
\caption{Physical finger-spread configurations. (A) Fully adducted. (B) Maximum spread. (C) Index, ring, and little fingers flexed at maximum spread. (D) V-shaped posture. (E) Reduced index--middle spacing with ring and little fingers flexed.}
\label{fig:spread_physical}
\end{figure*}

On the physical prototype, the index, ring, and little fingers remain fully flexion-capable at maximum spread (Fig.~\ref{fig:spread_physical}(C)); the middle finger can subsequently close as well, with small lateral compliance between the brick-built fingers accommodating the final closure. Finger spread can also be adjusted while selected fingers remain flexed, transitioning from the V-shaped posture in Fig.~\ref{fig:spread_physical}(D) to the narrower index--middle spacing in Fig.~\ref{fig:spread_physical}(E).

\section{Control, Teleoperation, and Simulation}
\begin{figure*}[!t]
\centering
\includegraphics[width=\textwidth]{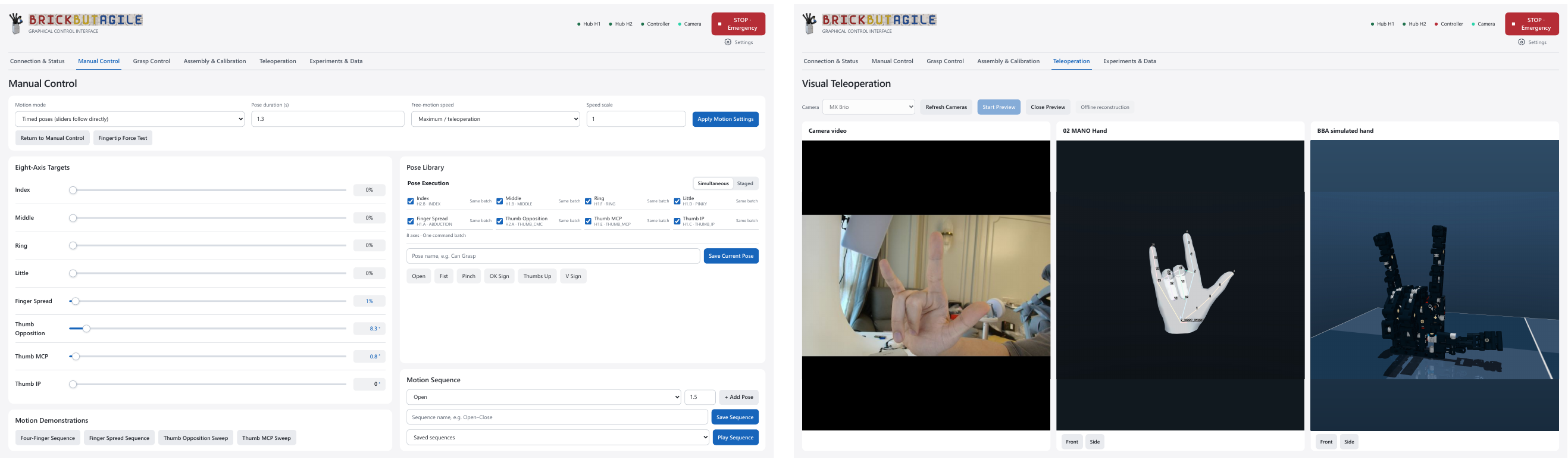}
\caption{Graphical control interface. Left: manual control and motion sequences. Right: visual teleoperation workspace displaying RGB camera input, 3D MANO hand reconstruction, and tendon-driven BBA simulation.}
\label{fig:control}
\end{figure*}
\subsection{Calibrated Control and Command Interface}
\label{sec:interaction}
Manual control, pose sequencing, and vision-based teleoperation share an eight-dimensional command vector (Fig.~\ref{fig:control}):
\begin{equation}
\mathbf u=[u_I,u_M,u_R,u_L,u_s,u_{T,C},u_{T,M},u_{T,I}]^\mathsf T.
\label{eq:command}
\end{equation}
Each component is normalized to $[0,1]$. The entries specify closure of the four non-thumb fingers, shared finger spread, and position targets for the thumb CMC, MCP, and IP joints. Calibration maps each normalized command to motor position using the channel reference, tendon winding direction, and usable travel.

Thumb CMC targets are converted to motor positions using the gear ratio, while MCP and IP targets use direction-specific calibration. Because MCP abduction and adduction change the IP tendon path length, the controller adjusts the IP motor target using an encoder-based estimate of MCP angle.

The host schedules target updates at 20~ms intervals and sends the latest targets to both SPIKE hubs over USB. Pybricks executes local motor-position control on the hubs~\cite{pybricks}. The graphical control interface provides per-channel speed, acceleration, and output limits, together with pose storage and motion sequencing. A Python API exposes the same calibrated controller for scripted execution and replay of interactively developed motions. Commands and encoder feedback are logged for replay and analysis.

\subsection{Vision-Based Teleoperation and Demonstration Capture}
\label{sec:teleop}
Vision-based teleoperation uses a single RGB camera and HaMeR~\cite{pavlakos2024} to reconstruct a 3D MANO hand mesh and local joint rotations~\cite{romero2017} from each detected hand crop, with temporal filtering on $\mathrm{SO}(3)$ suppressing monocular tracking jitter. Four reference poses---flat adducted, full fist, maximum finger spread, and thumb opposition---calibrate the operator's finger-flexion, spread, and thumb-opposition ranges, mapping scale-normalized finger curl, pairwise inter-finger spread angles, and 3D thumb posture into the command vector $\mathbf u$ for real-time control. During teleoperation, synchronized RGB video, MANO poses, commands, and encoder feedback are recorded for demonstration replay and analysis.

\subsection{Tendon-Driven Hand Simulation}
BBA provides URDF and MuJoCo XML models derived from the editable Studio assembly. In the physical hand, finger motion is not prescribed joint by joint, but emerges from tendon routing, passive PIP--DIP coupling, elastic return, and contact. To preserve these mechanics in simulation, we explicitly model the tendon-driven transmission in MuJoCo~\cite{todorov2012}. Four finger-flexion cables and one shared spread cable reproduce the active non-thumb transmission, while passive PIP--DIP coupling cables and elastic return bands represent the underactuated mechanics. The three thumb channels use the calibrated CMC, MCP, and IP mappings from Section~\ref{sec:interaction}. Cable paths update with mechanism motion, and the return bands use a measured effective stiffness of $27.45$~N/m. The simulator accepts the same eight motor commands as the physical hand.

\section{Experimental Evaluation}
\label{sec:experiments}
We evaluate whether a standard-LEGO hand with eight-input actuation can achieve repeatable fingertip motion, practical grasping and payload forces, full taxonomic grasp coverage, and reliable vision-based teleoperation.

\renewcommand{\emptyvalue}{---}

\subsection{Unidirectional Fingertip Repeatability}
\label{sec:motionforce}
\label{sec:repeatability}
\begin{figure}[!tbp]
\centering
\includegraphics[width=\columnwidth]{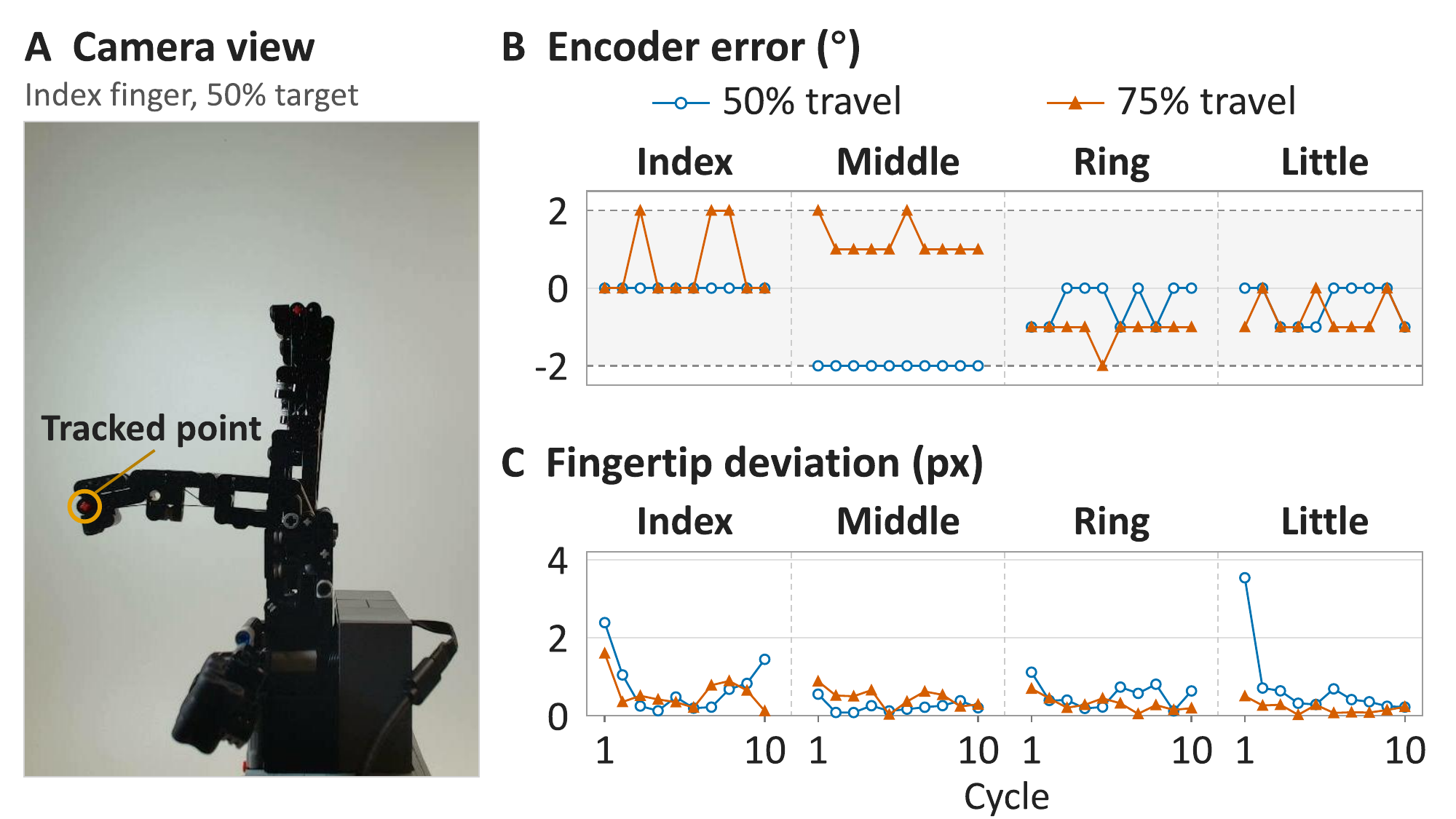}
\caption{Unidirectional repeatability of the four non-thumb fingers. (A) Example side-view recording; the circle marks the tracked point on the red axle of the index finger. (B) Motor encoder position errors across cycles; horizontal dashed lines indicate the $\pm2^\circ$ tolerance. (C) Distance $d_k$ of each tracked point from the mean position for its target. Circles and triangles denote 50\% and 75\% motor travel, respectively, with 10 cycles per target. Lines connect consecutive cycles.}
\label{fig:repeatability}
\end{figure}
Each non-thumb finger completes 10 cycles from open to 50\% motor travel, then to 75\% travel, and back to open, with finger spread fixed in the adducted configuration and the other fingers stationary. Recording both targets yields 80 endpoint measurements across the four fingers. At each settled target, we record the encoder reading and track the red-axle point at the fingertip from a fixed side view (Fig.~\ref{fig:repeatability}). Fingertip repeatability is quantified as the root-mean-square distance of the tracked points from their mean, computed separately for each finger and target.

All 80 encoder position errors fall within the $\pm2^\circ$ tolerance. Across the eight finger--target conditions, the root-mean-square fingertip positional scatter ranges from 0.24 to 1.21~px. Together with the $\pm2^\circ$ encoder bounds, this sub-pixel to single-pixel scatter confirms repeatable tendon-driven finger closure across cycles.

\subsection{Force Output and Static Payload}
\label{sec:forcepayload}

\begin{figure}[!tbp]
\noindent\makebox[\columnwidth][c]{\includegraphics[width=0.65\columnwidth]{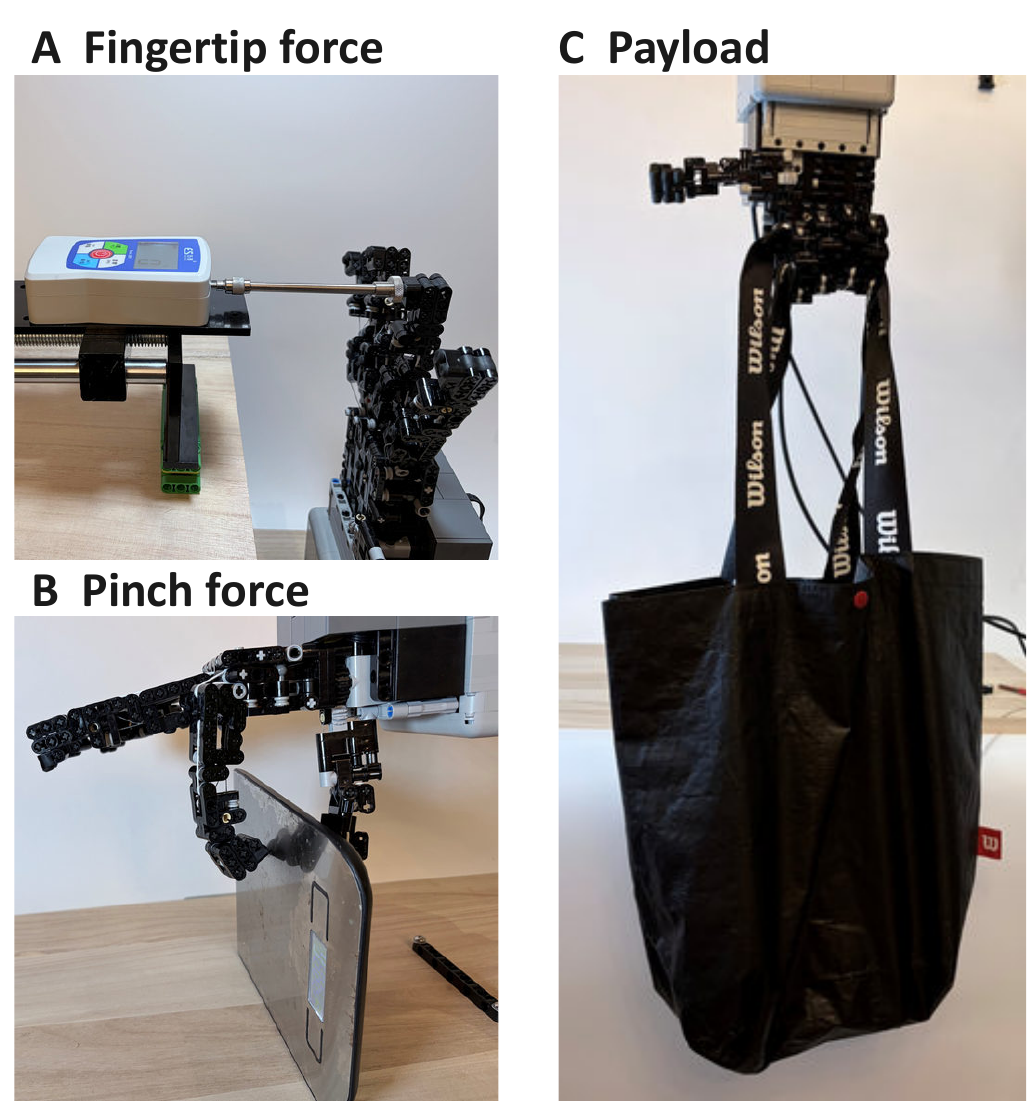}}
\caption{Force and payload test setups. (A) Fingertip force. (B) Thumb--index pinch force. (C) Bag-strap payload test assessing load-induced opening of the MCP joints.}
\label{fig:force}
\end{figure}

\begin{figure*}[!t]
\centering
\includegraphics[width=0.90\textwidth]{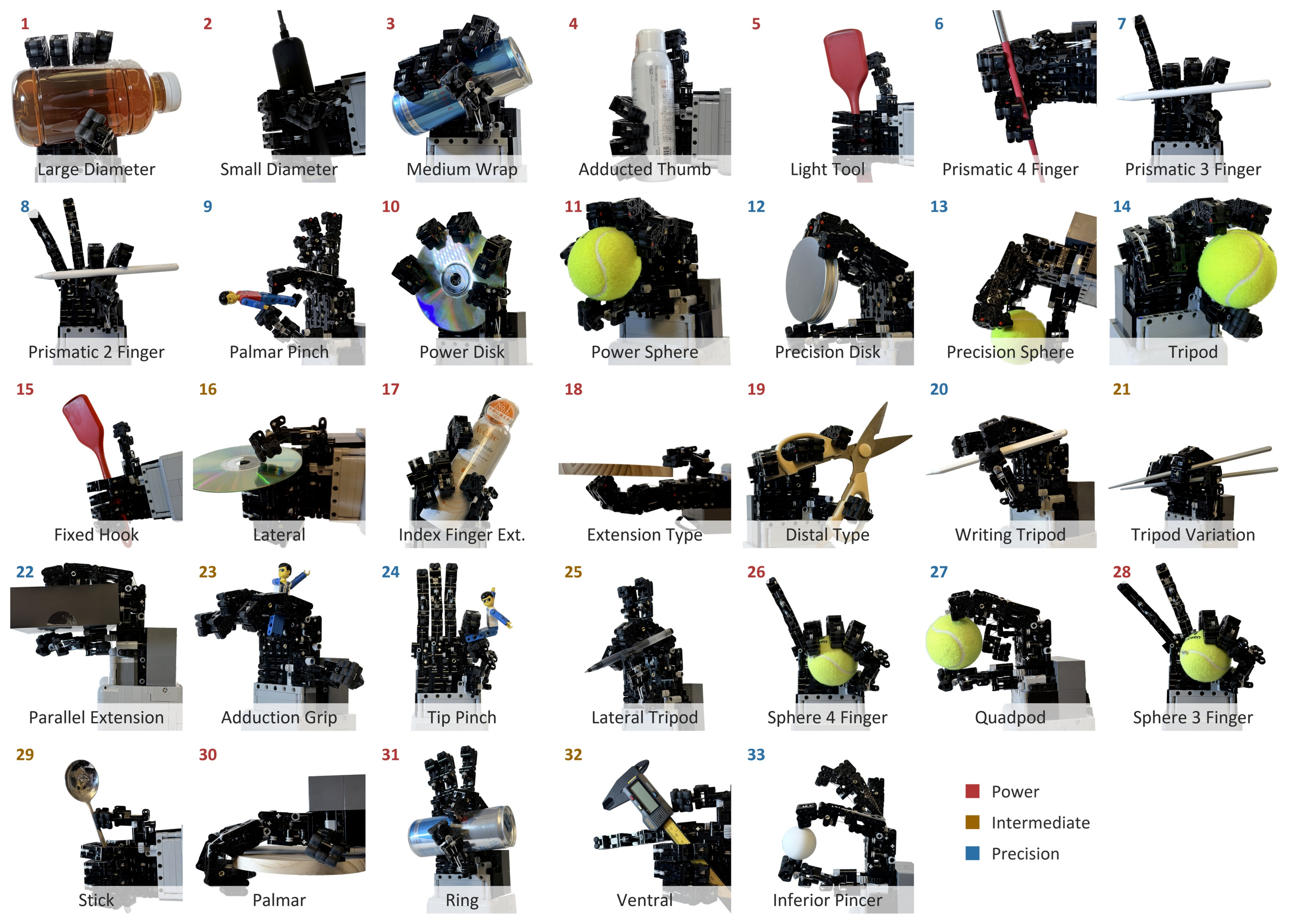}
\caption{BBA reproduces all 33 grasp types in the GRASP taxonomy~\cite{feix2016}, shown in numerical order with their original names. Number colors distinguish power, intermediate, and precision grasps according to the original classification.}
\label{fig:taxonomy}
\end{figure*}

\subsubsection{Fingertip Force}
With the hand base fixed, each non-thumb finger is commanded to flex against a stationary force gauge along its measurement axis (Fig.~\ref{fig:force}(A)). The mean maximum fingertip force across the four fingers is 3.30~N, with individual maxima of 3.8, 3.3, 3.0, and 3.1~N for the index, middle, ring, and little fingers, respectively.

\subsubsection{Pinch Force}
The thumb and index finger compress a force-measuring plate (Fig.~\ref{fig:force}(B)). Across three trials, the maximum pinch force is 2.94~N.

\subsubsection{Static Payload}
Static payload is evaluated with the four fingers supporting a weighted bag by its straps (Fig.~\ref{fig:force}(C)), using $15^\circ$ of load-induced MCP opening as the limit, a threshold also used in~\cite{zorin2025ruka,lin2026craft}. Without glue or threaded fasteners, the interlocking LEGO structure supports 2.35~kg below this limit---comparable to the 2.2~kg reported for Allegro Xella under the same test---while exceeding $15^\circ$ at 3.0~kg due to elastic deflection across the brick connections.

\subsection{Grasp Coverage and Thumb Opposition}
\label{sec:graspcoverage}
\subsubsection{GRASP Taxonomy}
BBA reproduces all 33 grasp types in the GRASP taxonomy~\cite{feix2016} on the physical hand (Fig.~\ref{fig:taxonomy}). These grasps include both enveloping configurations and localized contacts on slender rods and thin plates; in particular, shared finger spread and three-axis thumb opposition enable inter-finger lateral grasps (Adduction Grip), wide-span spherical grasps (Sphere 3 Finger and Sphere 4 Finger), and multi-finger tool grasps (Writing Tripod and Tripod Variation).

\subsubsection{Kapandji Test}
\begin{figure}[!tbp]
\centering
\includegraphics[width=\columnwidth]{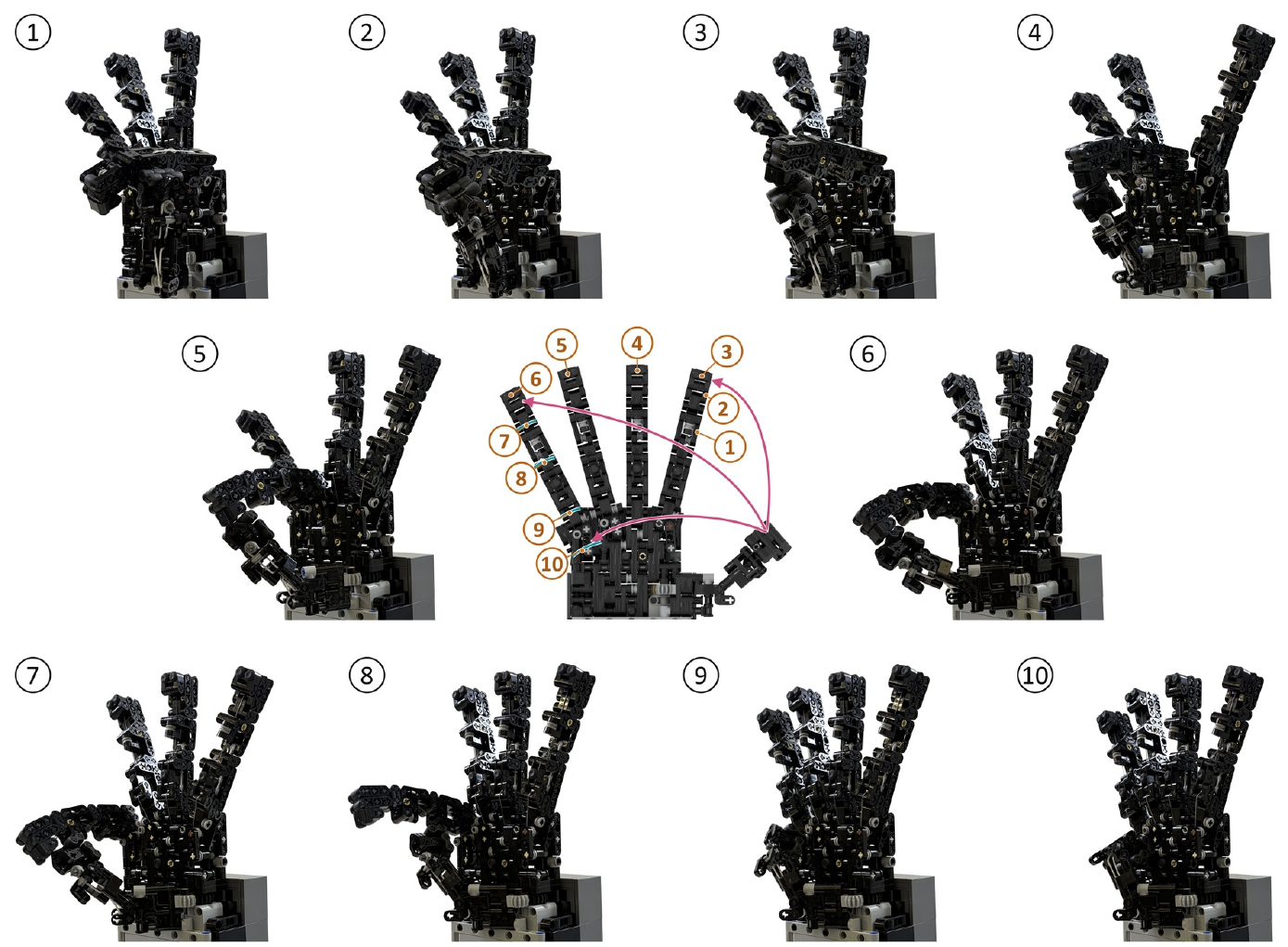}
\caption{Kapandji test. The central schematic identifies targets 1--10, and the surrounding photographs show the corresponding physical poses. Magenta arrows indicate the direction of thumb opposition; short cyan arcs mark the corresponding crease locations.}
\label{fig:kapandji}
\end{figure}

The Kapandji test~\cite{kapandji1986} assesses thumb opposition across the fingers and palm. BBA reaches all ten targets, scoring 10/10 (Fig.~\ref{fig:kapandji}).

\subsubsection{Thumb--Finger Contact Regions}
\begin{figure}[!tbp]
\centering
\includegraphics[trim=0 27.75bp 0 0,clip,width=0.70\linewidth]{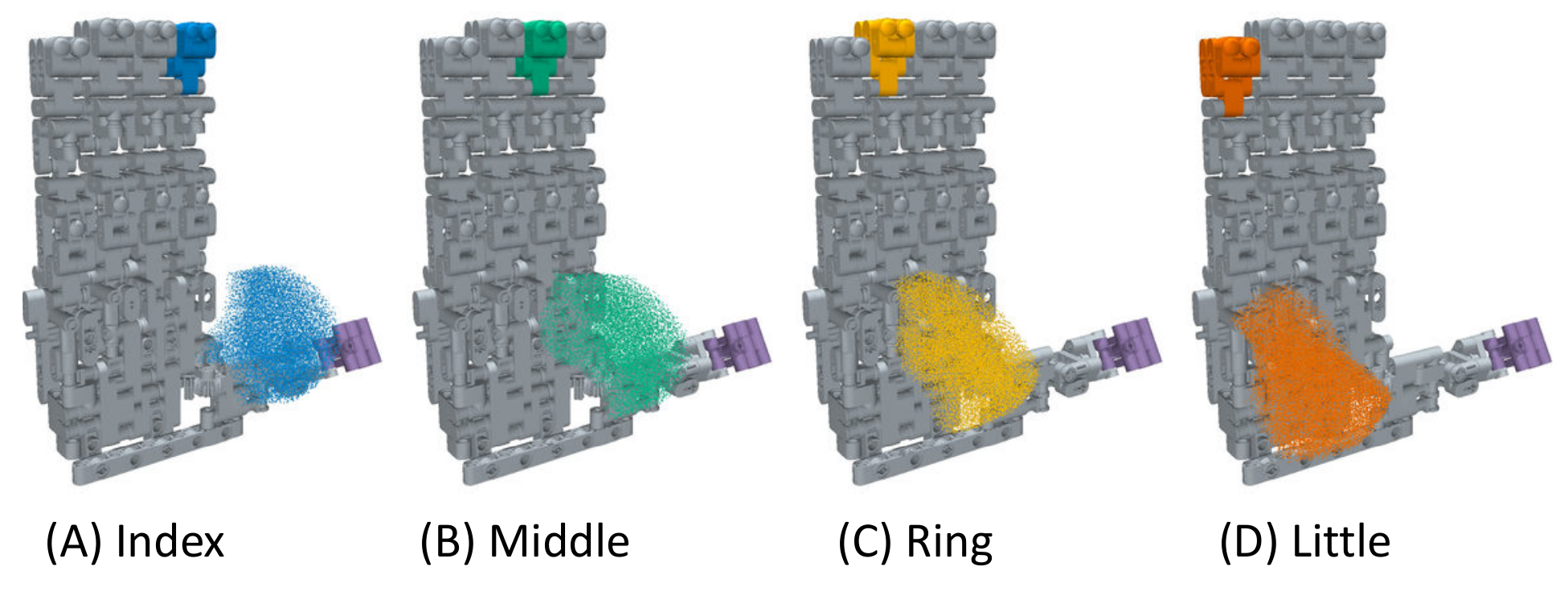}
\caption{Regions of near contact between the thumb and non-thumb fingertip surfaces in the digital model. Colors identify the finger opposing the thumb.}
\label{fig:thumb_opposition_surface}
\end{figure}

For each thumb--finger pair, we sample 250,000 joint configurations to estimate near-contact regions in the digital model. After excluding fingertip interpenetration, we retain positions where the minimum distance between the full fingertip surfaces is at most 1~mm (Fig.~\ref{fig:thumb_opposition_surface}), confirming opposition regions across all four fingertips.

\begin{figure}[!tbp]
\centering
\begingroup
\setlength{\tabcolsep}{1pt}
\begin{tabular*}{\columnwidth}{@{\extracolsep{\fill}}cccc@{}}
{\scriptsize\textbf{A} Can} &
{\scriptsize\textbf{B} Two-finger ball} &
{\scriptsize\textbf{C} Power ball} &
{\scriptsize\textbf{D} LEGO pinch} \\
\includegraphics[width=0.235\columnwidth]{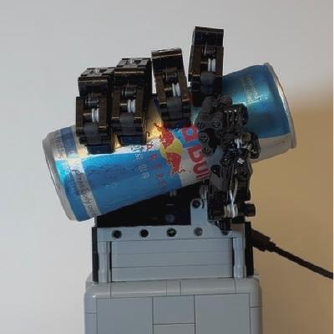} &
\includegraphics[width=0.235\columnwidth]{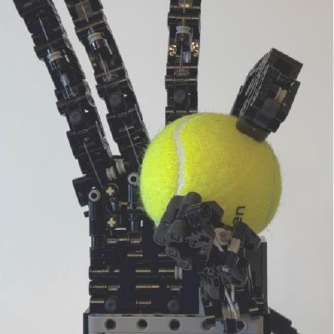} &
\includegraphics[width=0.235\columnwidth]{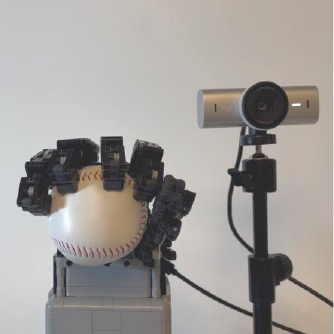} &
\includegraphics[width=0.235\columnwidth]{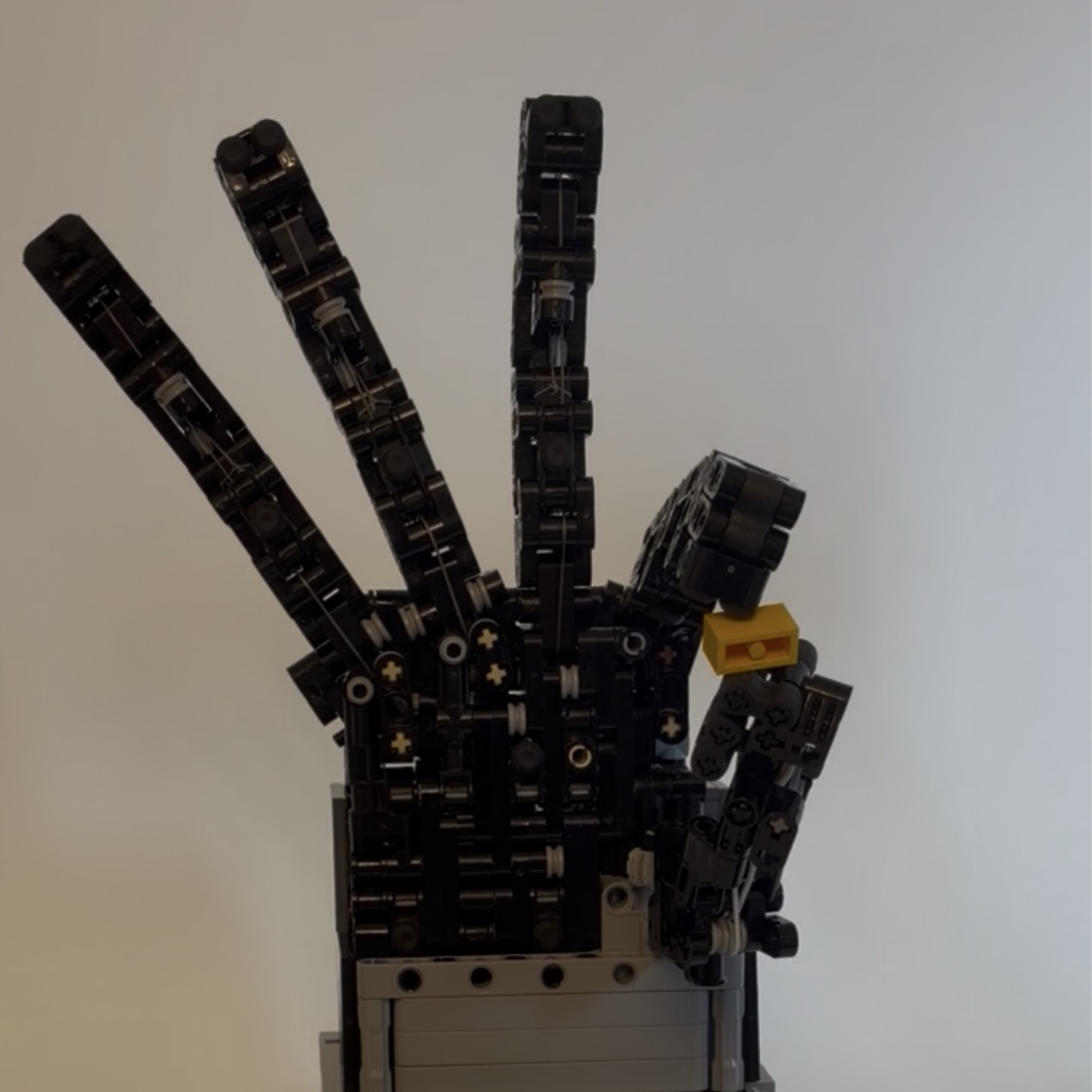}
\end{tabular*}
\endgroup
\caption{Vision-based teleoperation tasks. (A) Can grasp. (B) Two-finger ball grasp. (C) Whole-hand power grasp of a ball. (D) Thumb--index fingertip pinch of a $1\times2$ LEGO brick.}
\label{fig:teleoperation}
\end{figure}

\subsection{Vision-Based Teleoperation}
\label{sec:teleopeval}
Four tasks evaluate grasp execution using the vision-based teleoperation system in Section~\ref{sec:teleop}: an enveloping grasp of a can, a two-finger ball grasp, a whole-hand power grasp of a ball, and a thumb--index fingertip pinch of a $1\times2$ LEGO brick (Fig.~\ref{fig:teleoperation}). To isolate hand-level control from arm motion, the hand base is fixed while the operator commands grasping and release through continuous hand motion rather than stored poses. In each of 10 independent trials per task, the object is initially supported until grasped; success requires holding it without assistance for 5~s and then actively releasing it.

All four tasks succeed in 10/10 trials. These results show that the eight-channel interface enables reliable grasping, unassisted holding, and controlled release across distinct contact configurations under continuous visual hand-pose input. The $1\times2$ LEGO-brick task further demonstrates repeatable thumb--index fingertip coordination on a small object.

\section{Design for Educational Use}
Instruction-guided LEGO assembly and interactive control make BBA a hands-on platform for studying robotic grasping. Standard connections allow users to inspect and modify the joints and transmissions they build. Through the graphical control interface (Section~\ref{sec:interaction}), they can define poses, compose motion sequences, and compare commanded motions with motor feedback and video observations. Comparing free and object-constrained finger closure, for example, reveals how tendon actuation, passive coupling, and contact jointly shape finger motion.

\section{Conclusion}
BBA combines standard, reversible LEGO construction with selective finger and thumb control in a human-scale five-finger hand. Eight forearm inputs coordinate motion about eighteen joint axes without custom-fabricated parts. The prototype reproduces all 33 grasp types in the GRASP taxonomy, scores 10/10 on the Kapandji test, produces a mean maximum fingertip force of 3.30~N across the four non-thumb fingers, and holds a 2.35~kg payload with less than $15^\circ$ MCP opening. All four vision-based teleoperation tasks succeed in 10/10 trials. Together with editable digital assemblies, graphical control software, and tendon-driven simulation, BBA provides a reproducible platform for dexterous-manipulation research and a build--modify--control--observe workflow for hands-on robotics.

\FloatBarrier
\bibliographystyle{IEEEtran}
\bibliography{references}
\end{document}

%% file: layout.tex
\newcommand{\artwork}[4]{%
  \IfFileExists{figures/#1.pdf}{%
    \begin{minipage}[c][#3][c]{#2}\centering
      \includegraphics[width=#2,height=#3,keepaspectratio]{figures/#1.pdf}%
    \end{minipage}%
  }{%
    {\setlength{\fboxsep}{0pt}\setlength{\fboxrule}{0.35pt}%
    \fcolorbox{black!35}{black!2}{%
      \begin{minipage}[c][\dimexpr#3-2\fboxrule\relax][c]{\dimexpr#2-2\fboxrule\relax}
      \centering\sffamily\small
      \parbox{0.91\linewidth}{\centering #4}\par\vspace{6pt}
      {\scriptsize Artwork placeholder\quad /\quad height #3}
      \end{minipage}}}%
  }%
}
\newcolumntype{C}[1]{>{\centering\arraybackslash}m{#1}}
\renewcommand{\arraystretch}{1.06}
\newcommand{\emptyvalue}{\phantom{$0.00\,\pm\,0.00$}}

\hypersetup{pdftitle={Brick But Agile: A Compact, Modular Anthropomorphic LEGO Dexterous Hand for Education and Research},pdfauthor={}}
\makeatletter
\newcommand{\widefigureheader}[5]{%
 \twocolumn[{%
  \begin{minipage}{\textwidth}\centering
   \artwork{#1}{\textwidth}{#2}{#3}%
   \def\@captype{figure}\caption{#4}\label{#5}%
  \end{minipage}\vspace{10pt}%
 }]%
}
\newenvironment{fixedfigure}{\par\noindent\begin{minipage}{\columnwidth}\def\@captype{figure}\centering}{\end{minipage}\par\vspace{9pt}}
\newenvironment{fixedtable}{\par\noindent\begin{minipage}{\columnwidth}\def\@captype{table}\centering}{\end{minipage}\par\vspace{9pt}}
\makeatother

\makeatletter
\newcommand{\bbateaser}[1]{%
  \def\@bbateasercontent{%
    \begin{minipage}{\textwidth}
      \centering
      \includegraphics[width=\textwidth]{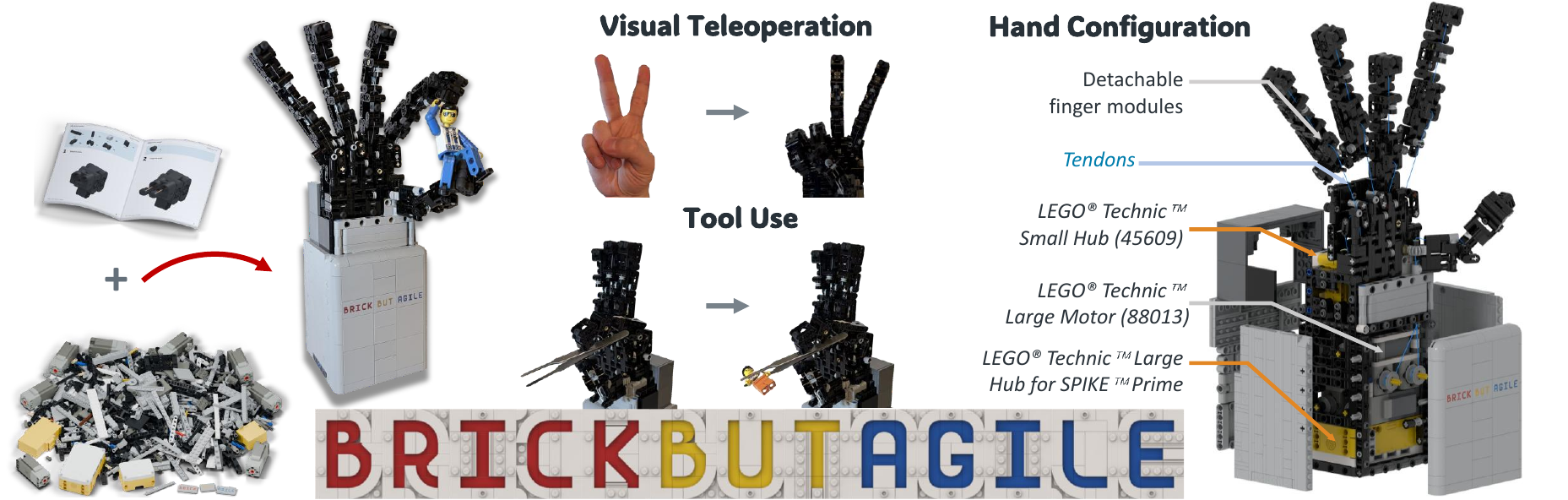}\par
      \def\@captype{figure}%
      \caption{#1}\label{fig:teaser}%
    \end{minipage}%
    \par\vspace{0.6\baselineskip}%
  }%
  \IEEEaftertitletext{\@bbateasercontent}%
}
\makeatother